# An Information Theoretic Approach to Quantify the Stability of Feature Selection and Ranking Algorithms

Rocío Alaiz-Rodríguez[a,*], Andrew C. Parnell[b]

[a]*Department of Electrical, Systems and Automatic Engineering, Universidad of León. Campus de Vegazana s/n, 24071 León, Spain*
[b]*Hamilton Institute, Maynooth University, Maynooth, Ireland*

## Abstract

Feature selection is a key step when dealing with high-dimensional data. In particular, these techniques simplify the process of knowledge discovery from the data in fields like biomedicine, bioinformatics, genetics or chemometrics by selecting the most relevant features out of the noisy, redundant and irrelevant features. A problem that arises in many of these applications is that the outcome of the feature selection algorithm is not stable. Thus, small variations in the data may yield very different feature rankings. Assessing the stability of these methods becomes an important issue in the previously mentioned situations, but it has been long overlooked in the literature. We propose an information-theoretic approach based on the Jensen-Shannon divergence to quantify this robustness. Unlike other stability measures, this metric is suitable for different algorithm outcomes: *full ranked* lists, *top-k* lists (feature subsets) as well as the lesser studied *partial ranked* lists that keep the k best ranked elements. This generalized metric quantifies the difference among a whole set of lists with the same size, following a probabilistic approach and being able to give more importance to the disagreements that appear at the top of the list. Moreover, it possesses desirable properties for a stability metric including correction for change, and upper/lower bounds and conditions for a deterministic selection. We illustrate the use of this stability metric with data generated in a fully controlled way and compare it with popular metrics including the Spearman's rank correlation and the

---

*Corresponding Author
*Email addresses:* rocio.alaiz@unileon.es (Rocío Alaiz-Rodríguez ), Andrew.Parnell@mu.ie (Andrew C. Parnell)

Kuncheva's index on feature ranking and selection outcomes respectively.

## 1. Introduction

Feature selection is a key step in many classification problems [22, 48, 5], in particular in those with high dimensional datasets. It is well known that the size of the training data set needed to calibrate a model grows exponentially with the number of dimensions (the curse of the dimensionality problem). Feature selection techniques measure the importance of the features according to the value of a given function [22]. The main motivation to implement these techniques has been to improve the classification performance by selecting an optimum subset of features. Numerous papers have examined feature selection with respect to classification performance [42, 12, 33].

Additionally, the process of knowledge discovery from the data in fields like biomedicine, bioinformatics, genetics or chemometrics is simplified with the use of feature selection methods. Removing the noisy and irrelevant features while keeping the most relevant features is essential for understanding the underlying process. In the medical field, for instance, it is well known that preventive screening for colorectal cancer and specialized care are changing the trends in reported mortality [3]. For these risk prediction applications, reducing the data dimensionality can mitigate overfitting and improve model performance [4, 13, 17, 14] to identify people at increased risk of developing this type of cancer. Moreover, feature selection methods become an important tool to uncover the risk factors (features, in the context of this work) for this disease [30, 16].

Identifying the most relevant features for the problem studied has been the goal of many research papers. It has been applied to discriminate different types of cancer [8, 21], to categorize healthy and diseased tissue [18], to identify people with higher risk to develop a disease [16] or to select genes related to a disease [43, 6, 2].

Although feature selection techniques are of great help to identify the most relevant features in these domains, a problem that arises in many practical problems is that the outcome of the feature selection algorithm does not tend to be stable in the sense that small variations in the data may yield to very different feature rankings. Stability (or robustness) issues have long been overlooked in the literature. However, the topic of robustness of feature selection techniques has attracted an increasing interest in the machine learning field in the past few years [29, 25, 11, 50, 23, 51, 40, 1, 20, 39]. The issues have arisen perhaps as a consequence of the difficulties of reproducing different research findings. Evaluating the stability of ranked feature (or top-k) lists that come out of the feature ranking (or selection) techniques becomes crucial before trying to gain insight into the data. Otherwise, the conclusions derived from the study may be completely unreliable. In order to measure the stability, suitable metrics for each output format (full ranked feature lists, partial ranked lists or top-k lists) of the feature selection algorithms are required.

The Spearman's rank correlation coefficient [28, 29?] and Canberra distance [25] have been proposed to measure the similarity when the outcome representation is a full ranked list. When the goal is to measure the similarity between top-k lists (also referred to as feature subsets), a wide variety of measures have been proposed: Jaccard distance [29?], an adaptation of the Tanimoto distance [29], Kuncheva's stability index [32], Consistency measures [45], Dice-sorense's index [36], Ochiai's index [52] or Percentage of overlapping features [24]. Among all of them, the Spearman rank correlation coefficient $S_R$, Jaccard stability index [29, 41] or Kuncheva's stability index [32] $KI$ are possibly the most widely used metrics.

An alternative that lies between full ranked lists (all features with ranking information) and partial lists (a subset with the top-k features, where all of them are assumed to have the same importance) is the use of partial ranked lists, that is, a list with the top-k features and the relative ranking among them. This approach has been used in the information retrieval domain [7] to evaluate queries and it seems more natural when the goal is to analyze a subset of features. To our knowledge only a modified version of the Canberra distance has been proposed for this purpose [27].

It seems reasonable that when it comes to assess the robustness of feature selection techniques, two ranked lists should be considered much less similar if their differences occurred at the "top" rather than at the "bottom" of the lists. Unlike metrics such as the Kendall's tau and the Spearman's rank correlation coefficient that do not capture this information, we propose a stability measure based on information theory that takes this into consideration. Our proposal is based on mapping each ranked list into a probability distribution and then, measuring the dissimilarity among these distributions using the information-theoretic Jensen-Shannon divergence. Furthermore, this single metric, $S_{JS}$ (Similarity based on the Jensen-Shannon divergence) applies to full ranked lists, partial ranked lists as well as top-k lists with equal length. Furthermore, it also fulfills the desirable properties for a stability metric.

The rest of this paper is organized as follows: In Section 2 we formulate the problem of feature selection. In Section 3 we describe the robustness issue and common approaches to deal with it. The new metric based on the Jensen-Shannon divergence $S_{JS}$ is presented in Section 4. A discussion of the desired properties of the stability metrics is in Section 5 and a comparison among several feature and ranking techniques in Section 6. Experimental evaluation is shown in Section 7. Finally Section 8 summarizes our main conclusions.

## 2. Feature Selection Techniques

Consider a training dataset $D = \{(\mathbf{x}_i, d_i), i = 1, \ldots, M\}$ with M examples and a class label $d$ associated with each sample. Each sample $\mathbf{x}_i$ is a $t$-dimensional vector $\mathbf{x}_i = (x_{i1}, x_{i2}, \ldots x_{it})$ where each component $x_{ij}$ represents the value of a given feature $f_j$ for that example $i$, that is, $f_j(\mathbf{x}_i) = x_{ij}$.

Feature selection techniques measure the importance of a feature or a subset of features according to a given measure. These techniques may provide many benefits, the most important ones being [48]: (a) to mitigate the curse of dimensionality, (b) to gain a deeper insight into the underlying processes that generated the data, and (c) to provide faster and more cost-effective prediction models.

From a structural point of view, these algorithms can be divided into three categories [22, 10, 12]: filter, wrapper and embedded approaches. The filter techniques rely on general characteristics of the training data to rank the features according to a metric without involving any learning algorithm. The wrapper approaches incorporate the interaction between the feature selection process and the classification model in order to determine the value of a given feature subset. Finally, in the embedded techniques, the feature search mechanism is built into the classifier model and are therefore specific to a given inductive learning algorithm.

From a functional point of view the output of a feature selection algorithm may be a ranking (weighting-score) on the features or feature set. Obviously, representation changes are possible and thus, a feature subset can be extracted from a full ranked list by selecting the most important features and a partial ranked list can be also derived directly from the full ranking by removing the least relevant features.

Consider now a feature ranking algorithm that leads to a ranking vector $\mathbf{r}$ with components

$$\mathbf{r} = (r_1, r_2, r_3, \ldots, r_t) \tag{1}$$

where $1 \leq r_i \leq t$. Note that 1 is considered the highest rank.

Consider also a top-k list as the outcome of a feature selection technique

$$\mathbf{s} = (s_1, s_2, s_3, \ldots, s_t), s_i \in \{0, 1\} \tag{2}$$

where 1 indicates the presence of a feature and 0 the absence and $\sum_{i=1}^{t} s_i = k$ for a top-k list.

Lists with a full ranking of features can be converted into top-k lists that contain the most important k features. Converting a ranking output into a feature subset is easily conducted according to

$$s_i = \begin{cases} 1 & \text{if } r_i \leq k \\ 0 & \text{if } otherwise \end{cases}$$

## 3. Related work: Robustness of feature selection techniques

A fundamental property of a feature selection method is its robustness [23, 39, 46]. This becomes critical in many domains where the stability of a feature selection method is crucial for interpretation by domain experts. Robustness has been defined as the sensitivity of the method to small perturbations in the training set [29].

Non-stability of feature selection is a problem that may appear in practical applications, but in particular it is more noticeable when the available dataset is small and the feature dimensionality is high, as is common in biomedicine, bioinformatics, and chemometrics. Instability issues make the feature rankings unreliable for clinical use. Therefore, it becomes essential to provide metrics to evaluate the robustness of given feature selection techniques when applied to our data. Efforts have also been made in order to increase the robustness of feature selection methods [40, 1, 35, 44, 9].

Suppose we ran a feature ranking algorithm $K$ times and obtained a set of rankings $\mathbf{A} = \{\mathbf{r}_1, \mathbf{r}_2, \ldots, \mathbf{r}_K\}$. For the purpose of illustration, Figure 1 shows an example where instances are defined by ten features ($t = 10$) and the feature ranking algorithm is applied to five different subsamples of the data ($K = 5$)

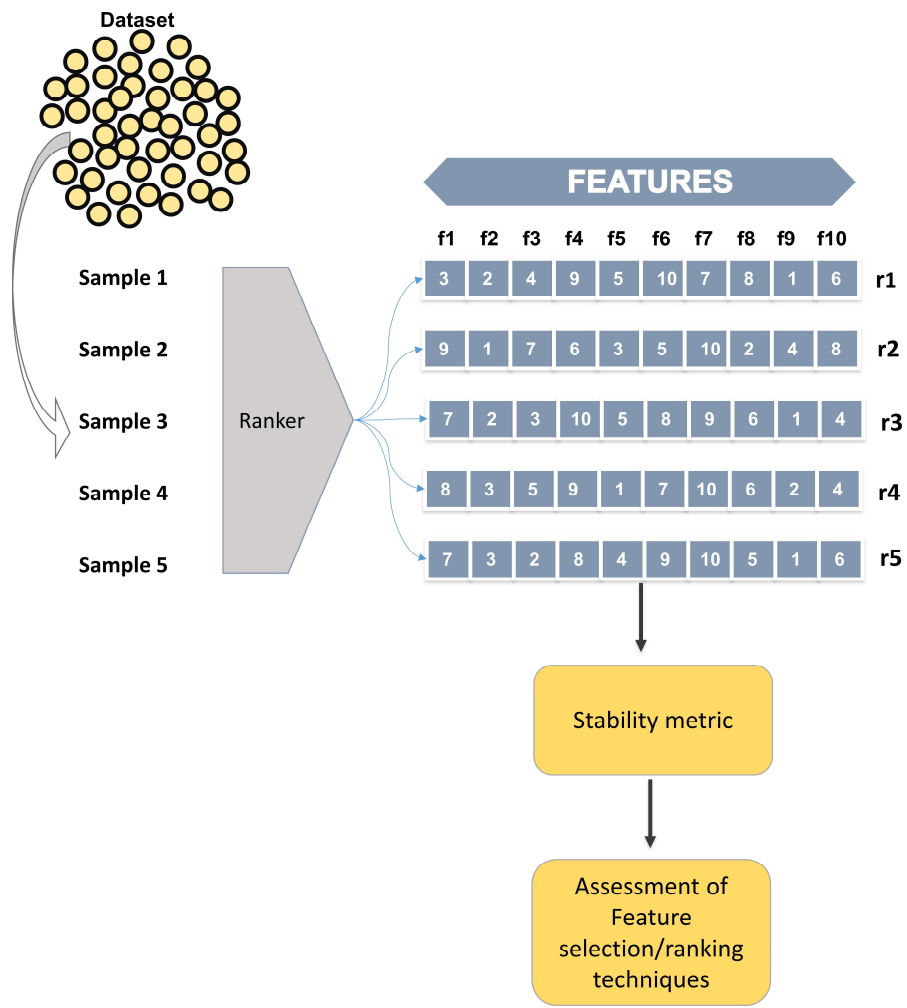

Figure 1: Illustration of the stability problem for feature ranking methods.

Once obtained, the dissimilarity among ranked lists can be measured at different levels:

- Among full ranked lists

- Among feature subsets (top-k lists)

- Among partial ranked lists (top-k ranked lists)

Thus, the outcomes for full ranked lists can be gathered in a matrix $\mathsf{A}$ with elements $r_{ij}$ with $i = 1, \ldots, t$ and $j = 1, \ldots, K$ that indicate the rank assigned in run $j$ for feature $i$. Note that $\mathsf{A}_{fr}$ stands for set of lists with full ranking.

$$
\mathsf{A}_{fr} = [\mathbf{r}_1^i \quad \mathbf{r}_2^i \quad \mathbf{r}_3^i \quad \mathbf{r}_4^i \quad \mathbf{r}_5^i] =
\begin{bmatrix}
3 & 9 & 7 & 8 & 7 \\
2 & 1 & 2 & 3 & 3 \\
4 & 7 & 3 & 5 & 2 \\
9 & 6 & 10 & 9 & 8 \\
5 & 3 & 5 & 1 & 4 \\
10 & 5 & 8 & 7 & 9 \\
7 & 10 & 9 & 10 & 10 \\
8 & 2 & 6 & 6 & 5 \\
1 & 4 & 1 & 2 & 1 \\
6 & 8 & 4 & 4 & 6
\end{bmatrix}_{10 \times 5}
$$

Figure 2 shows the top-4 ranked lists and the top-4 lists for the example presented above. Additionally, some of the stability metrics that are commonly applied for each output format are also shown.

The outcomes for the top-4 lists can also be gathered in a matrix $\mathsf{A}_f$ with elements $s_{ij}$ with $i = 1, \ldots, t$ and $j = 1, \ldots, K$ that indicate whether or not the feature-$i$ has been selected among the top-4 most relevant in the run-$j$.

$$
\mathsf{A}_s = [\mathbf{s}_1^i \quad \mathbf{s}_2^i \quad \mathbf{s}_3^i \quad \mathbf{s}_4^i \quad \mathbf{s}_5^i] =
\begin{bmatrix}
1 & 0 & 0 & 0 & 0 \\
1 & 1 & 1 & 1 & 1 \\
1 & 0 & 1 & 0 & 1 \\
0 & 0 & 0 & 0 & 0 \\
0 & 1 & 0 & 1 & 1 \\
0 & 0 & 0 & 0 & 0 \\
0 & 0 & 0 & 0 & 0 \\
0 & 1 & 0 & 0 & 0 \\
1 & 0 & 1 & 1 & 1 \\
0 & 0 & 1 & 1 & 0
\end{bmatrix}_{10 \times 5}
$$

In the case of dealing with partial ranked lists, the set of lists can be

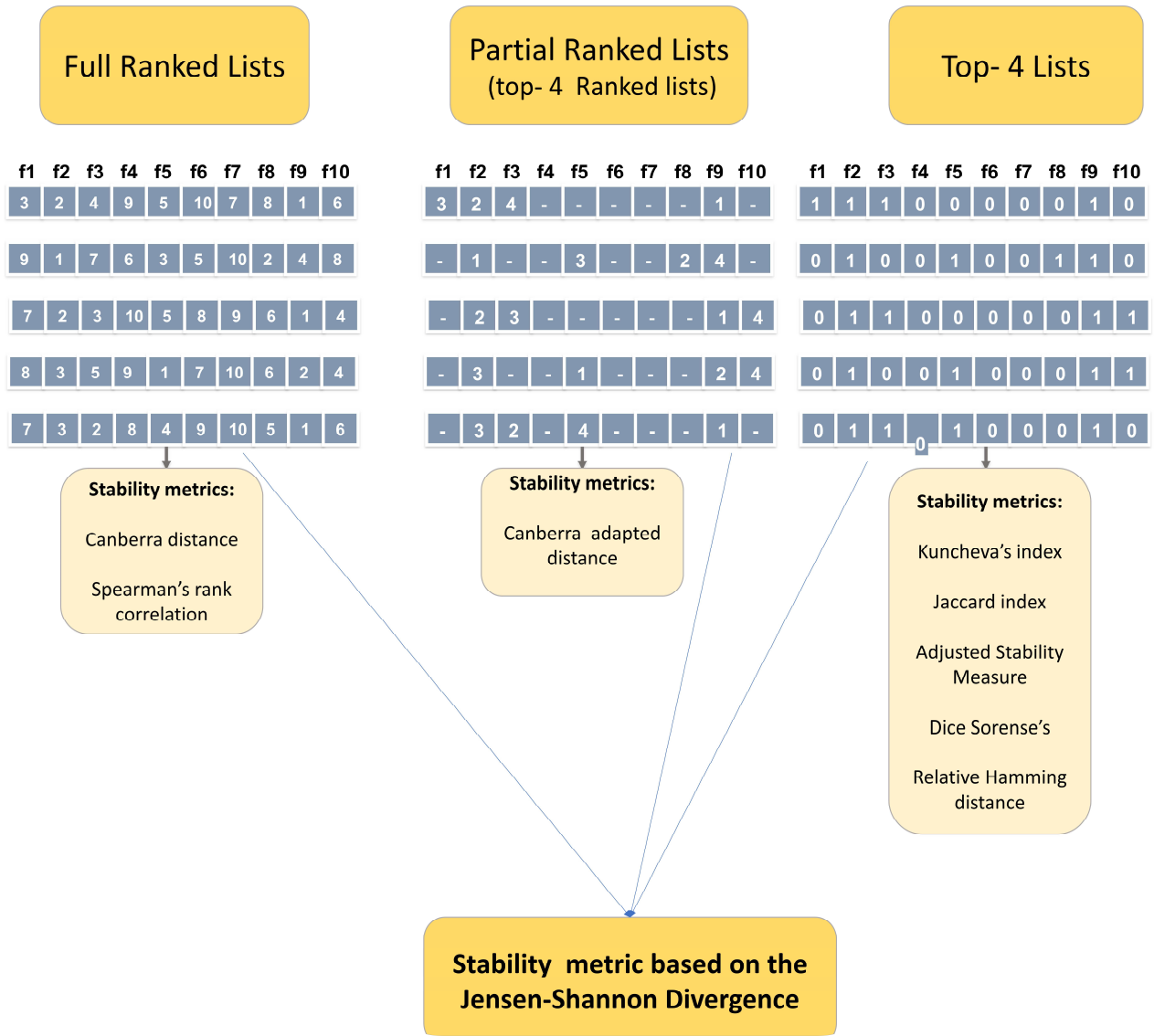

Figure 2: Formats for feature selection techniques: full ranked lists, partial ranked lists (top-k ranked lists) and top-k lists.

represented in matrix $\mathbf{A}_{pr}$

$$\mathbf{A}_{pr} = \begin{matrix} 3 & 0 & 0 & 0 & 0 \\ 2 & 1 & 2 & 3 & 3 \\ 4 & 0 & 3 & 0 & 2 \\ 0 & 0 & 0 & 0 & 0 \\ 0 & 3 & 0 & 1 & 4 \\ 0 & 0 & 0 & 0 \\ 0 & 0 & 0 & 0 & 0 \\ 0 & 0 & 2 & 0 & 0 & 0 \\ 1 & 4 & 1 & 2 & 1 \\ 0 & 0 & 4 & 4 & 0 \end{matrix}_{10 \times 5}$$

Ideally, this metric should be bounded by constants that do not depend on the size of the sublist $k$ or on the total number of features $t$. Additionally, it should have a constant value for randomly generated subsets/rankings.

In general, stability is quantified as follows: (a) Given a set of rankings (or subsets), pairwise similarities are computed and then reduced to a single metric by averaging. (b) Defining a function applied on matrix $\mathbf{A}$ but not based on pairwise similarities. (c) Visual analysis of stability.

### 3.1. Robustness analysis based on computing pairwise similarities

The most widely use approach to evaluate the stability of a feature selection (or ranking) algorithm that provides several results $\mathbf{A} = \{ \mathbf{r}_1, \mathbf{r}_2, \ldots \mathbf{r}_K \}$, is to compute pairwise similarities and average the results. This approach leads to a scalar value:

$$\Phi(\mathbf{A}) = \frac{2}{K(K-1)} \sum_{i=1}^{K-1} \sum_{j=i+1}^{K} S_M(\mathbf{r}_i, \mathbf{r}_j) \tag{3}$$

where $S_M$ refers to any similarity metric which takes as input the appropriate format of $\mathbf{A}$.

### 3.1.1. Similarity metric for full ranked lists

Consider $\mathbf{r}$ and $\mathbf{r}^i$ the output of a feature ranking technique applied to two subsamples of $\mathbf{D}$. The Spearman's rank correlation coefficient [29, ?, 37, 24] and Kendall's tau coefficient [47, 49] have been proposed to measure the similarity between rankings. Of the two, Spearman's rank correlation coefficient ($S_R$) is perhpas the most popular. The $S_R$ between two ranked lists **r** and **r**$^j$ is defined by

$$S_R(\mathbf{r}, \mathbf{r}^j) = 1 - 6 \sum_{i=1}^{t} \frac{(r_i - r_i^j)^2}{t(t^2 - 1)} \tag{4}$$

where $r_i$ is the rank of feature-$i$ and $t$ the total number of features. $S_R$ values range from -1 to 1. It takes the value one when the rankings are identical and the value zero when there is no agreement between rankings. This metric is only suitable for lists with the same size.

### 3.1.2. Similarity metric for feature subsets

When the goal is to measure the similarity between feature subsets (also referred as top-k lists) different authors have proposed similarity metrics: Jaccard distance [29], Tanimoto distance [29], Kuncheva's stability index [32], Relative Hamming distance [19], Consistency measures, Dice-sorense's index [36], Ochiai's index or Percentage of overlapping features [24]. Of these, the Kuncheva's stability index and the Jaccard distance appear to be the most widely accepted [32, 1, 24].

Let consider now **s** and **s**$^j$ as the output vector of a feature selection algorithm applied to two different subsamples of $D$. The Kuncheva's index ($KI$) for these two top-k lists is given by

$$KI(\mathbf{s}, \mathbf{s}^j) = \frac{ot - k^2}{k(t - k)} \tag{5}$$

where $t$ is the total number of features, $o$ is the number of features that are present in both lists and $k$ is the length of the sublists, that is, $\sum_{t} s_i = \sum_{t} s_i^j = k$. The KI satisfies $-1 < KI \leq 1$, achieving its maximum when the two lists are identical ($o = k$) and values close to zero for independently drawn lists **s** and **s**$^j$ (i.e. $o$ expected to be around $k^2/t$).

The Jaccard stability index (JI) is defined as

$$JI(\mathbf{s}, \mathbf{s}^j) = \frac{|\mathbf{s} \wedge \mathbf{s}^j|}{|\mathbf{s} \vee \mathbf{s}^j|} = \frac{o}{l} \tag{6}$$

where **s** and **s**$^j$ are the two feature subsets, $o$ is the number of features that are common in both lists and $l$ the number of features that appear only in one of the two lists. The JI lies in the range (0, 1).

### 3.1.3. *Similarity metrics for top-k ranked lists*

These include metrics like the Canberra distance, initially proposed to assess the similarity between full feature rankings. These were extended to partial ranked lists using a location parameter [26]. Additionally the Pearson's rank correlation coefficient [29**?** , 37, 24] can also fall in this category.

### 3.2. *Robustness analysis based on a function definition*

Generally, we can define a function $\Phi(A)$ to avoid computing all pairwise similarities. A popular measure in this category is the Relative Weighted Consistency Measure CWrel [45]. This stability metric is a direct function of the frequency of the features after feature selection. Other proposals within this category include the frequency of selection normalized by the number of feature subsets and averaged over all features [20].

### 3.3. *Visual Analysis of Robustness*

The outcome of a feature ranking algorithm can be interpreted as a point in a high dimensional space (with $t$ dimensions). The stability of a pairwise ranking can be viewed as computing distances between points in that high dimensional space and averaging the results. These (scalar) metrics can be seen as projections to one dimensional space and their use only provides guidance as to where the feature selector stands in relation to a stable and a random ranking algorithm.

The use of graphical methods as a simple alternative approach to evaluate the stability of feature ranking algorithms has been proposed in [16]. It has been highlighted that if we change from a projection to a space with one dimension, into a space with two or more dimensions, we can conduct a visual analysis that allows the user to visually assess stability as well as establish comparisons with other feature ranking or selection methods.

In [16], a dimensionality reduction technique like Multi-Dimensional Scaling (MDS) [15] has been proposed for a visual analysis of robustness. It allows the projection of data from a high dimensional space to a 2D or 3D space while preserving the distance in the original high dimensional space.

Figure 3 illustrates this approach with several feature ranking algorithms: *FR-a*, *FR-b*, *FR-c*, *FR-d*, *FR-e*. The algorithms are run on seven sub-samples of the data. This figure allows the user to see in a single figure that the most unstable algorithm is *FR-a* since the points are very scattered. The outcomes of *FR-d*, however, are clustered together. The same applies to the *FR-e* and these therefore are the most stable. This figure also allows the user to see that *FR-e* generates a similar ranking to *FR-d*. Finally, note that *FR-c* is very different to the aforementioned groups.

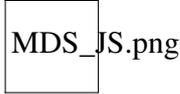

Figure 3: Visual-based stability analysis for five hypothetical feature rankings (FR).

## 4. An Information Theoretic Approach to Measure Robustness

We propose a stability measure based on the Jensen-Shannon Divergence [34] able to measure the discrepancy among several full ranked lists, among partial ranked lists (top-k ranked lists) and also among feature subsets (top-k lists). When the ranking is taken into account, the differences at the top of the list would be considered more important than differences at the bottom part, regardless of whether it is a full or a partial list. When we focus on top-k lists, all the features would be given the same importance. This stability metric applies to feature sets or feature rankings with the same cardinality.

Our approach to measure the stability of feature selection/ranking techniques is based on mapping the output of the feature selection/ranking algorithm into a probability distribution. Then, the distance between these distributions is measured with the Jensen-Shannon divergence [34]. Below we present our proposal for full ranked lists and then Section 4.1 and Section 4.2 describe its extension to top-k ranked lists and top-k lists, respectively.

Given the output of a feature ranking algorithm, features at the top of the list should be given the highest probability (or weight) and it should smoothly decrease according to the rank. Thus, following [7] the ranking vector $\mathbf{r} = (r_1, r_2, r_3, \ldots, r_t)$ would be mapped into the probability vector $\mathbf{p} = (p_1, p_2, p_3, \ldots, p_t)$ where

$$p_i = \frac{1}{2t} \left( 1 + \sum_{j=0}^{t-n} (r_i + j)^{-1} \right) \tag{7}$$

where by design $\sum_{i=1}^{t} p_i = 1$. We can thus quantify the similarity between two ranked lists $\mathbf{r}$ and $\mathbf{r}^i$ by measuring the divergence between the distributions $\mathbf{p}$ and $\mathbf{p}^i$ associated with them.

The most widely used metric for measuring the difference between two probability distributions is the Kullback-Leibler (KL) divergence $D_{KL}$ [31], given by

$$D_{KL}(\mathbf{p}||\mathbf{p}^i)) = \sum_i p_i \log \frac{p_i}{p^i_i} \tag{8}$$

This measure is always non negative, taking values from 0 to $\infty$, and $D_{KL}(p||q)) = 0$ if $p = q$. The KL divergence, however, has two important drawbacks, since (a) in general it is asymmetric ($D_{KL}(p||q) \neq D_{KL}(q||p)$) thus not a true distance measure, and (b) it does not generalize to more than two distributions. For this reason, we use the related Jensen-Shannon divergence [34], that is a symmetric version of the Kullback-Leibler divergence and is given by

$$D_{JS}(\mathbf{p}||\mathbf{p}^i) = \frac{1}{2}(D_{KL}(\mathbf{p}||\bar{\mathbf{p}}) + D_{KL}(\mathbf{p}^i||\bar{\mathbf{p}})) \tag{9}$$

where $\bar{\mathbf{p}}$ is the average of the distributions.

Given a set of $K$ distributions $\mathbf{p}_1, \mathbf{p}_2, \ldots, \mathbf{p}_K$, where each one corresponds to a run of a given feature ranking algorithm, we can use the Jensen-Shannon divergence to measure the similarity among the distributions produced by different runs of the feature ranking algorithm, what can be expressed as

$$D_{JS}(\mathbf{p}_1, \ldots, \mathbf{p}_K) = \frac{1}{K} \sum_{i=1}^K D_{KL}(\mathbf{p}_i||\bar{\mathbf{p}}) \tag{10}$$

or equivalently as

$$D_{JS}(\mathbf{p}_1, \ldots, \mathbf{p}_K) = \frac{1}{K} \sum_{j=1}^K \sum_{i=1}^t p_{ij} \log \frac{p_{ij}}{\bar{p}} \tag{11}$$

with $p_{ij}$ being the probability assigned to feature i in the ranking output j and $\bar{p}_i$ the average probability assigned to feature i.

Some desirable constraints that this stability measure possesses includes:

· It falls in the interval [0 ,1]

· It takes the value zero for completely random rankings

· It takes the value one for stable rankings

· It is invariant to the ordering of the ranking probability distributions

We define the stability metric $S_{JS}$ (Stability based on the Jensen-Shannon divergence) as:

$$S_{JS}(\mathbf{p}_1, \ldots, \mathbf{p}_K) = 1 - \frac{D_{JS}(\mathbf{p}_1, \ldots, \mathbf{p}_K)}{D^*_{JS}(\mathbf{p}_1, \ldots, \mathbf{p}_K)} \tag{12}$$

where $D_{JS}$ is the Jensen-Shannon Divergence among the $K$ ranking outcomes and $D^*_{JS}$ is the divergence value for a ranking generation that is completely random. In a random setting, $\bar{p}_i = 1/t$ which leads to a constant value $D^*_{JS}$

$$D^*_{JS}(\mathbf{p}_1, \ldots, \mathbf{p}_K) = \frac{1}{K} \sum_{j=1}^{K} \sum_{i=1}^{t} p_{ij} \log(p_{ij} t) = \frac{1}{K} K \sum_{i=1}^{t} p_i \log(p_i t) = \sum_{i=1}^{t} p_i \log(p_i t) \tag{13}$$

where $p_i$ is the probability assigned to a feature with rank $r_i$. Note that this maximum value depends exclusively on the number of features and it can be computed beforehand with the mapping provided by (7).

We can check that:

- For a completely stable ranking algorithm, $p_{ij} = \bar{p}_i$ in (11). That is, the rank of feature-$j$ is the same in any run-$i$ of the feature ranking algorithm. This leads to $D_{JS} = 0$ and a stability metric $S_{JS} = 1$

- A random ranking will lead to $D_{JS} = D^*_{JS}$ and therefore $S_{JS} = 0$

- For any ranking neither completely stable nor completely random, the similarity metric $S_{JS} \in (0, 1)$. The closer to 1, the more stable the algorithm is.

### 4.1. Extension to partial ranked lists

The similarity between partial ranked lists, that is, partial lists that contain the top-k features with relative ranking information can be also measured with the $S_{JS}$ metric. In this case, the probability is assigned to the top-k ranked features is:

$$p_i = \begin{cases} \frac{1}{2k}\left(1 + \sum_{j=0}^{k-n}(r_i + j)^{-1}\right) & \text{if } r_i \leq k \\ 0 & \text{otherwise} \end{cases} \tag{14}$$

The $S_{JS}$ is computed according to (12) with the normalizing factor $D^*_{JS}$ given by (13) and the probability $p_i$ assigned to a feature with rank $r_i$ computed as stated in (14).

## 4.2. Extension to feature subsets

When we deal with feature subsets with a given number of top-k features, a uniform probability is assigned to the selected features according to

$$p_i = \begin{cases} \dfrac{1}{k} & \text{if } r_i \leq k \\ 0 & \text{otherwise} \end{cases} \tag{15}$$

The $S_{JS}$ is computed according to (12) with the probability $p_i$ assigned to a feature according to (15) and the normalizing factor $D_{JS}^*$ given by

$$D_{JS}^*(\mathbf{p}_1, \ldots, \mathbf{p}_K) = \sum_{i=1}^{t} p_i \log(p_i t) = \sum_{i=1}^{t} \frac{1}{k} \log\left(\frac{1}{k} t\right) = \log\left(\frac{t}{k}\right) \tag{16}$$

where $k$ is the length of the sublist and $t$ the total number of features.

## 5. Properties of Stability Metrics

There are some properties that a stability metric should possess so that it allows for a useful interpretation of stability and similarly comparisons among feature selection/ranking techniques. Kuncheva [32] was the first to provide a list of desirable properties for a similarity measure $S_M$ between two feature subsets of equal length (top-k lists). We should keep in mind that the $S_M$ is then averaged over all pairs to obtain a stability metric according to (3). We list the various properties in this section.

These properties, however, only refer to distance metrics and and they do not necessarily imply that the stability index obtained by computing pairwise similarities have the same properties as the stability measure. On the other hand, there are other proposals such as [45] (or ours) that are not based on calculating pairwise similarities. These properties were later refined in [39] where the authors study the properties from the wider viewpoint of the stability metric and not for the similarity metric.

Nogueira et al. [39] focused on feature selection techniques that may select feature subsets of arbitrary cardinality identifying some properties necessary for a given stability measure. These desirable properties are: upper and lower bounds, correction for chance, maximum stability, and fully defined. They further showed that many stability measures widely used in the literature do not possess all these properties.

The four properties proposed by [32] are:

### Property 1: Upper and Lower Bounds

The stability metric $\Phi$ should have upper and lower bounds that do not depend on the total number of features or the feature subset length.

### Property 2: Maximum ⟷ Deterministic Selection

The stability metric $\Phi(\mathbf{A})$ should reach its maximum if-and-only-if all feature sets in $\mathbf{A}$ are identical.

### Property 3: Correction For Chance

When the selection is random, that is feature sets of size $k_i$ have an equal probability of being drawn, the expected value of $\Phi(\mathbf{A})$ should be constant, which is set for convenience to 0.

### Property 4: Fully Defined

The stability metric $\Phi(\mathbf{A})$ should be completely defined for any set of features. This property ensures the stability metric can cope with feature subsets of any size.

This latter property enables the application of a stability metric to domains where the feature selection algorithm may return subsets with different number of features. Nonetheless, we consider this property as optional but not essential since there are many scenarios in which the number of selected features is fixed to a number $k$ for a given study.

## 5.1. Properties of the Stability Metric $S_{JS}$

The $S_{JS}$ stability measure presented in this work possesses the first three of the aforementioned properties:

### Property 1: Upper and Lower Bounds

The stability metric $S_{JS}$ takes values in the interval [0 ,1]

### Property 2: Maximum ⟷ Deterministic Selection

The stability metric $S_{JS}$ reaches its maximum value 1 if-and-only-if all feature sets in $\mathbf{A}$ are identical.

### Property 3: Correction For Chance

When the selection is random, there is a normalizing term $D^{\star}_{JS}(\mathbf{p}_1, \ldots, \mathbf{p}_K)$ that corresponds to the divergence value for a feature set or ranking that is completely random. In that case, $S_{JS}$ takes the value 0.

The stability metric we propose in this paper focusses on problems where the feature rankings or feature subsets have the same length, hence the *fully defined* property does not apply.

## 6. Comparison of Stability Metrics

The output of a technique that selects the most relevant features may come in the form of: (a) a full ranking of features , (b) a top-k ranked list either with a fixed length $k$ or different cardinality, (c) feature sets with a fixed length or (d) feature sets with arbitrary length. Table 6 summarizes for which output format some widely known stability metrics can be applied.

It is evident that the stability metrics developed for feature subsets cannot deal with rankings and in general the opposite is also true. There are metrics, however, such us the Canberra distance initially proposed for full feature rankings. By using the location parameter, this distance can also be computed between upper partial lists of the original rankings [26]. Later, it was extended [27] to partial ranked lists that may have different length.

Many of the metrics proposed for computing the distance between feature sets can cope with lists of different length: Tanimoto, ASM, CWrel, the relative Hamming distance, the Jaccard distance and Dice-Sorense's index. The popular Kuncheva index applies only to lists with equal length, though. By contrast, our stability measure $S_{JS}$ can deal with full and partial ranked lists as well as feature sets. In all cases the feature lists should have the same number of elements.

Table 6 shows a general overview of the previously mentioned three properties: Upper and Lower Bounds, Maximum Deterministic Selection and Correction For Chance. Note that the Fully defined property defined in [39] was included in Table 6 (last column with label top-k lists with arbitrary length). As before, we consider it useful to determine whether or not a stability metric can be applied to a given output format but in our opinion this cannot be viewed as an essential property by itself.

Some well known stability metrics do not verify the *correction for change* property, such as Tanimoto, Relative Hamming distance, Jaccard distance, Dice-sorense's index or the Relative Weighted Consistency CWrel. Other stability metrics like the Adjusted Stability Measure (ASM) or the Relative Weighted Consistency CWrel do not fulfill the *Maximum* property. The main strength of the metric $S_{JS}$ is that it can deal either with feature rankings

Table 1: Eligible stability metrics for different feature rankings and feature subset formats

| Stability metric | Full ranked lists | Partial ranked lists | Partial ranked lists with different length | Feature subset lists | Feature subset lists with different length |
|---|---|---|---|---|---|
| Canberra distance [26] | Yes | Yes | - | - | - |
| Canberra adapted distance [27] | Yes | Yes | Yes | - | - |
| Spearman's rank correlation coefficient [29] | Yes | Yes | - | Yes | - |
| Tanimoto [29] | - | - | - | Yes | Yes |
| Adjusted Stability Measure ASM [38] | - | - | - | Yes | Yes |
| Kuncheva's stability index [32] | - | - | - | Yes | - |
| Relative Weighted Consistency CWrel [45] | - | - | - | Yes | Yes |
| Relative Hamming distance [19] | - | - | - | Yes | Yes |
| Jaccard distance [29] | - | - | - | Yes | Yes |
| Dice-sorense's index [36] | - | - | - | Yes | Yes |
| Our proposal: Jensen-Shannon stability metric | Yes | Yes | - | Yes | - |

(full /partial) or feature subsets and additionally, it possesses the essential properties for a stability metric.

Table 2: Properties of stability metrics for feature selection methods

| Stability metric | Upper and Lower bounds | Maximum | Correction |
|---|---|---|---|
| Tanimoto [29] | Yes | Yes | - |
| Adjusted Stability Measure ASM [38] | Yes | - | Yes |
| Kuncheva's stability index [32] | Yes | Yes | Yes |
| Relative Hamming distance [19] | Yes | Yes | - |
| Jaccard distance [29] | Yes | Yes | - |
| Dice-sorense's index [36] | Yes | Yes | - |
| Relative Weighted Consistency CWrel [45] | Yes | - | - |
| Our proposal: Jensen-Shannon stability | Yes | Yes | Yes |

## 7. Experimental Results

In this section we illustrate the stability metric $S_{JS}$ for the outcomes of some hypothetical feature ranking algorithms. We generate sets of $N = 100$ rankings of $l = 2000$ features. We simulate several Feature Ranking (FR) algorithms:

- FR-0 with 100 random rankings, that is, a completely random FR algorithm

- FR-1 with one fixed output, and 99 random rankings.

- FR-2 with two identical fixed outputs, and 98 random rankings.

- FR-$i$ with $i$ identical fixed outputs, and $100 - i$ random rankings.

- FR-100 with 100 identical rankings, that is, a stable FR technique.

Figure 4 shows our stability metric based on the Jensen-Shannon divergence ($S_{JS}$) compared ton the Spearman's rank correlation coefficient ($S_R$) for FR techniques that vary from completely random (FR-0, on the left) to completely stable (FR-100 on the right). For the FR-0 method, the stability metric $S_{JS}$ takes the value 0, while its value is 1 for the stable FR-100 algorithm. Note that $S_{JS}$ takes similar values to the Spearman's rank correlation coefficient $S_R$.

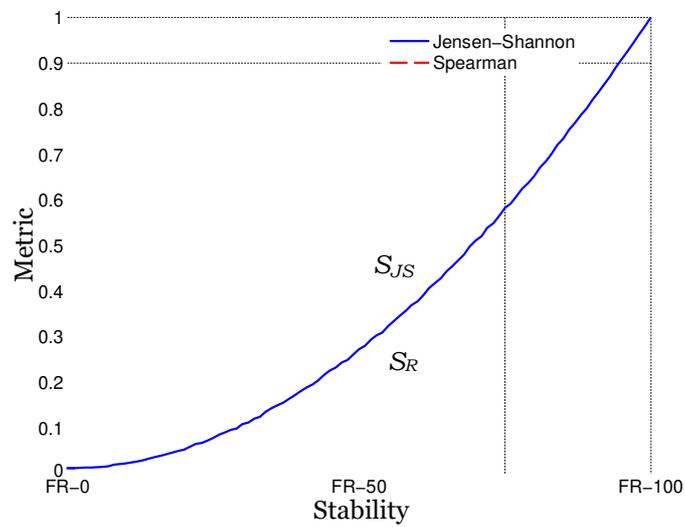

Figure 4: $S_{JS}$ metric and Spearman rank correlation for Feature Ranking (FR) techniques that vary from completely random (FR-0 on the left) to completely stable (FR-100 on the right).

Suppose now we have some Feature Selection (FS) techniques, for which stability needs to be assessed. These FS methods (FS-0,FS-1,...,FS-100) have been obtained from the corresponding FR techniques described above, extracting the top-k features ($k = 600$). In the same way, they vary smoothly from a completely random FS algorithm (FS-0) to stable FS a completely stable one (FS-100). The Jensen-Shannon metric $S_{JS}$ together with the Kuncheva Index (KI) are depicted for top-600 lists in Figure 5. Note that the $S_{JS}$ metric applied to top-k lists provides similar values to the KI metric. The Jensen-Shannon based measure $S_{JS}$ can be applied to full ranked lists and partial lists, while the KI is only suitable for partial lists and the $S_R$ only to full ranked lists.

Generating partial ranked feature lists is an intermediate step between:
(a) generating and comparing full ranked feature lists that are, in general, very long and (b) extracting sublists with the top-k features, but with no relevance information for each feature. The $S_{JS}$ metric based on the Jensen-Shannon divergence also allows to compare these partial ranked lists.

Suppose we have sets of sublists with the 600 most important features out of 2000 features. We generated several sets of lists: some of them show high differences in the lowest ranked features whilst others show high differences in the highest rank features. The same sublist can come either with the ranking information (partial ranked lists) or with no information about the feature importance (top-k lists). The overlap among the lists is around 350 features. Figure 6 shows the value $S_{JS}$ (partial ranked lists), $S_{JS}$ (top-k list) and the Kuncheva index (top-k lists) for the lists.

Even though the lists have the same average overlap (350 features), some of them show more discrepancy about which are the top features (Figure 6, on the right), while other sets show more differences at the bottom of the list. The KI can not handle this information since it only works with top-k lists and therefore, it assigns the same value for these very different situations. When the $S_{JS}$ works at this level (top-k list), it also gives the same measure for all the scenarios. The $S_{JS}$ can also handle the information provided in partial ranked lists, considering the importance of the features and therefore assigning a lower stability value for those sets of lists with high differences at the top of the lists, that is with high discrepancy about the most important features. Likewise, it assigns a higher stability value for those sets where the differences appear in the least important features, but there is more agreement about the most important features. Figure 6 illustrates this fact where $S_{JS}$ (for partial ranked lists) varies according to the location of the differences in the list, while $S_{JS}$ (top-k lists) and the KI assign the same value regardless of where the discrepancies appear.

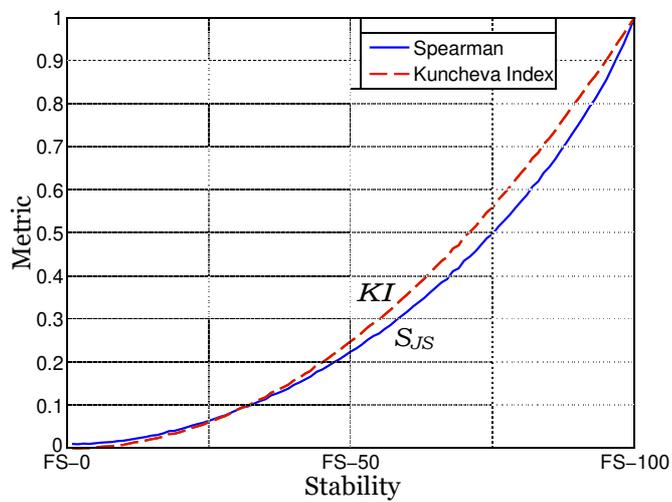

Figure 5: $S_{JS}$ metric and the KI for Feature Selection (FS) techniques that vary from completely random (FS-0 on the left) to completely stable (FS-100 on the right). The metrics work on top-k lists with k=600.

Next, consider the situation where the most important 600 features out of 2000 have been extracted and the overlap among the top-600 lists is 100%. We have evaluated several scenarios:

- The feature ranks are identical in all the lists (Identical)

- The ranking of a given feature is assigned randomly (Random)

- Neither completely random nor completely identical.

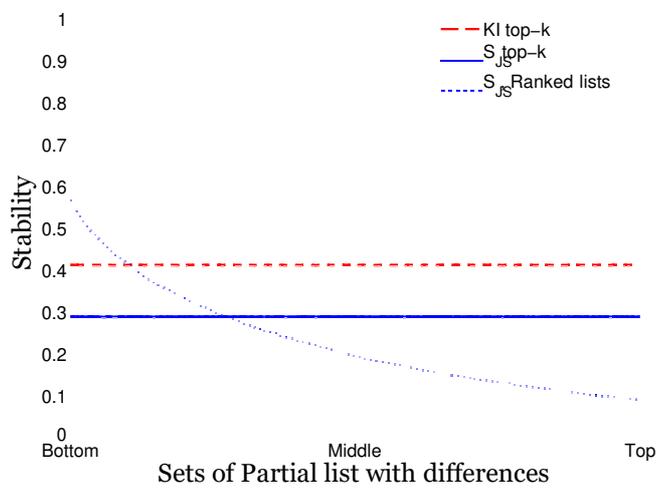

Figure 6: $S_{JS}$ (partial ranked lists), $S_{JS}$ (top-k list) and the Kuncheva index (top-k lists) for Feature Selection (FS) techniques that extract the top-600 features out of 2000. The overlap among the lists is around 350 common features. The situations vary smoothly from sets of partial lists with differences at the bottom of the list (left) to sets of lists that show high differences at the top of the list (right).

Working with top-k lists (KI), the stability metrics provide a value of 1 that is somewhat misleading considering the different scenarios that may appear. It seems natural that, even though all agree about the 600 most important features, the stability metric should be lower than 1 when there is low agreement about which are the most important features. The $S_{JS}$ measure allows us to work with partially ranked lists and therefore establish differences between these scenarios. Figure 7 shows the $S_{JS}$ (partial ranked lists) and the $S_{JS}$, KI (top-k lists) highlights this fact. $S_{JS}$ (partial ranked lists) takes a value slightly higher than 0.90 for a situation where there is complete agreement about which are the most important 600 features, but complete discrepancy about their importance. Its value increases to 1 as the randomness in the feature ranking assignment decreases. In contrast with this, *KI* would assign a value of 1 which may mislead when studying the stability issue.

## 8. Conclusions

Quantifying the stability of feature selection/ranking algorithms becomes a crucial issue when the aim of these techniques is to gain insight into the underlying process. The stability of feature selection algorithms concerns a wide area of recent interest that includes the development of more robust feature selection techniques and different approaches to measure their stability.

In this work, we addressed the problem of assessing the stability and have proposed an information theoretic metric based on the Jensen-Shannon divergence ($S_{JS}$) able to capture the mismatch among the lists generated in different runs by a feature selection algorithm. From a functional point of view the output of a feature selection algorithm may be: a ranking of the features or a feature set. Unlike most metrics that are specifically designed for a given output format, this stability metric applies to: (i) full ranked feature lists, (ii) top-k features, that is to say, lists that contain the k most relevant features giving a uniform relevance to all them and (iii) partial ranked lists that keep the $k$ most ranked features. To our knowledge, no metric has been proposed so far which is able to measure the similarity at all these levels.

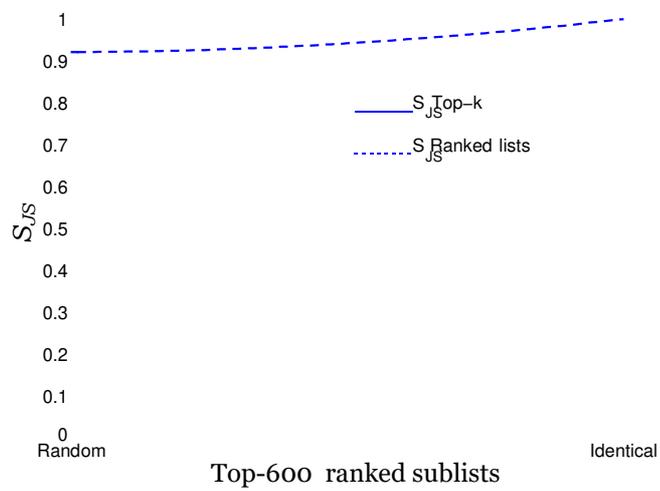

Figure 7: $S_{JS}$ (top-k list) and $S_{JS}$ (partial ranked lists) for Feature Selection (FS) techniques that extract the top-600 features out of 2000. The overlap among the sublists with 600 features is complete. The ranking assigned to each feature varies from FS techniques for which it is random (left) to FS techniques for which each feature ranking is identical in each sublist (right).

Unlike other metrics that evaluate pairwise similarities, $S_{JS}$ evaluates the whole set of lists directly (with the same size). Besides accepting whatever representation of the feature selection output, its behavior is: (a) close to the Spearman's rank correlation coefficient for full ranked lists and (b) similar to the Kuncheva's index for top-k lists. When the ranking is taken into account, the differences at the top of the list would be considered more important than differences that appear at the bottom part. Additionally, the new metric $S_{JS}$ quantifies the relative amount of randomness of the ranking/selection algorithm and when dealing with sublists, it is independent of the fixed number of features.

It is noteworthy that the $S_{JS}$ stability metric verifies the desired properties for a stability metric: upper and lower bounds, conditions for a deterministic selection and correction for change. Therefore, it enables a useful interpretation of stability as well as comparisons among feature selection/ranking techniques.

Potential future work includes the exploration of visual techniques with this new metric embedded and the extension of it to partial lists (either ranked or not) with different number of features.

## Acknowledgements

## Conflict of interest